# Differential Evolution for Efficient AUV Path Planning in Time Variant Uncertain Underwater Environment

Somaiyeh Mahmoud.Zadeh, David M.W Powers, Amir Mehdi Yazdani, Karl Sammut, Adham Atyabi

**Abstract** The Autonomous Underwater Vehicle (AUV) three-dimension path planning in complex turbulent underwater environment is investigated in this research, in which static current map data and uncertain static-moving time variant obstacles are taken into account. Robustness of AUVs path planning to this strong variability is known as a complex NP-hard problem and is considered a critical issue to ensure vehicles safe deployment. Efficient evolutionary techniques have substantial potential of handling NP hard complexity of path planning problem as more powerful and fast algorithms among other approaches for mentioned problem. For the purpose of this research Differential Evolution (DE) technique is conducted to solve the AUV path planning problem in a realistic underwater environment. The path planners designed in this paper are capable of extracting feasible areas of a real map to determine the allowed spaces for deployment, where coastal area, islands, static/dynamic obstacles and ocean current is taken into account and provides the efficient path with a small computation time. The results obtained from analyze of experimental demonstrate the inherent robustness and drastic efficiency of the proposed scheme in enhancement of the vehicles path planning capability in coping undesired current, using useful current flow, and avoid colliding collision boundaries in a real-time manner. The proposed approach is also flexible and strictly respects to vehicle's kinematic constraints resisting current instabilities.

## 1 Introduction

Robust path planning is an important characteristic of autonomy. It represents the possibilities for vehicles deployment during a mission in presence of different disturbances, which is a complicated procedure due to existence of the inaccurate, unknown, and varying information. Currents variations can strongly affect vehicle's motion by enforcing the vehicle to an undesired direction. Robustness of vehicle path planning enables an AUV to cope with adverse currents as well as exploit desirable currents to enhance the operation speed that results in considerable energy saving. To achieve path planning objectives, several approaches have been introduced, which could be classified based on two disciplines of deterministic and heuristic approach.

A plethora of research suggested deterministic methods for solving unmanned vehicle's path planning problem (Tam et al. 2009; Yilmaz et al. 2008; Kruger et al. 2007; Soulignac 2011), while path planning based deterministic methods on is carried out on repeating a set of predefined steps that search for the best fitted solution to the objectives (Tam et al. 2009). A mixed integer linear programming (MILP) is conducted Yilmaz et al. (2008) for multiple AUVs path planning problem. A non-linear least squares optimization technique is employed for AUV path planning through the Hudson River (Kruger et al. 2007) and sliding wave front expansion algorithm is applied to generate appropriate path for AUVs in presence of strong current fields (Soulignac 2011). Spatio-temporal wave front method (Thompson et al. 2010) and sliding wave front expansion methods (Soulignac et al. 2008; Soulignac 2011) is investigated on the same problem, respectively. Level set methods is applied by Lolla et al. (2012) for AUV time/energy optimum path planning taking current map into account. The "heuristic-grid search" approaches are another alternatives in dealing with NP-hard and multi-objective optimization problems. Carroll et al. (1992) represented the operation space with quad trees and applied A* to obtain efficient paths with minimum cost. An energy efficient trajectory planning based on A* search algorithm is proposed by Garau et al. (2009) taking quasi-static ocean current information into account, in which a common situation on actual AUVs are considered and thrust power is assumed to be constant.

[1]S.M. Zadeh, [1]D.M.W Powers, [1]A.M. Yazdani, [2]K.Sammut, [3]A. Atyabi

[1] School of Computer Science, Engineering and Mathematics, Flinders University, Adelaide, SA, Australia
[2] Centre for Maritime Engineering, Control and Imaging, Flinders University, Adelaide, SA, Australia
[3] Seattle Children's Research Institute, University of Washington, Washington, United States
e-mail: somaiyeh.mahmoudzadeh@flinders.edu.au
e-mail: david.powers@flinders.edu.au
e-mail: amirmehdi.yazdani@flinders.edu.au
e-mail: karl.sammut@flinders.edu.au
e-mail: adham.atyabi@seattlechildrens.org
S.M. Zadeh
School of Computer Science, Engineering and Mathematics, Flinders University, Adelaide, SA, Australia

A modified A* algorithm called "Constant-Time Surfacing A*" is employed by Fernandez-Perdomo et al. (2010) to provide a time optimal path for a glider using Regional Ocean Model (ROM) data. Later on, Pereira et al. (2011) applied A* to find a path with minimum risk for gliders according to historical data from the automated information system. A*-based path planer also used by Koay and Chitre (2013) to find path with minimum energy consumption considering variable ocean current. However, deterministic and heuristic methods like A* suffer from expensive computational cost and criticized for their weak performance in high-dimensional problems. The Meta heuristic algorithms propose robust solution in a very fast computational speed, in which the proposed solution will not necessarily correspond to the optimal solution (M.Zadeh et al. 2015; M.Zadeh et al. 2016). The importance of the ocean current in saving energy was emphasized by Alvarez et al. (2004), in which a GA based path planning was investigated for finding a safe path with minimum energy cost in a variable ocean characterized by strong currents, while avoiding getting trapped by local minima. An energy efficient path planner based on particle swarm optimization is studied by Witt and Dunbabin (2008) taking time-varying ocean currents into account. Ant colony asexual reproduction optimization is conducted by Asl et al. (2014) for path planning of mobile robots.

To satisfy the addressed challenges with aforementioned methods and handling different real-world challenges associated with ocean variability and uncertainty, this paper contributes an efficient Differential Evolution based (DE) based path planer that guarantees a successful operation of the vehicle.

For the purpose of this study, kinematic of the AUV, static current map and uncertain obstacles along with real map date are taken into account for promoting the capability of the path planner in handling real-world underwater variations. The implementation has been performed in MATLAB®2016, which includes the robustness assessment in terms of satisfying specified objectives and constraints of the AUV path planning problem. The organization of the paper is as follows. In Section 2, the path planning problem, kinematics of the vehicle and environmental modelling are stated. Section 3 discusses about application of Differential Evolution on AUV path planning problem and optimization criterion. The simulation results and evaluation of the proposed DE-based path planner is provided in Section 4. And, the Section 5 concludes the paper.

**2 Path Planning Considering Vehicle Kinematics and Ocean Dynamics**

The path planning is an NP-hard optimization problem in which the main goal is to minimize the travel time/distance, avoiding colliding obstacle(s), and coping current variations over the time. Efficient modeling of the operational terrain including offline map, dynamic/static time variant current, and different moving and static uncertain obstacles provides test bed for evaluating the path planner in situations that closely match the real world underwater environment.

To model a realistic marine environment, a three dimensional terrain $\Gamma_{3D}:\{2.5\ km^2\ (x\text{-}y),\ 1000\ m(z)\}$ is considered based on real map of the Benoit's Cove (located in Newfoundland, Canada), in which the terrain is covered by fixed and uncertain dynamic obstacles. In this research, information of static 2D turbulent current field has been applied, as the deep ocean current fields take long time to update and do not change immediately. The current dynamics $V_C=(u_c,v_c)$ is estimated and modelled using superposition of multiple viscous Lamb vortices and 2-D Navier-Stokes equation (Garau et al. 2006). The map is set with the size of 250×250 pixels where each pixel contains a current vector from the current map and corresponds to the area of $10m^2$. A k-means clustering method is applied to cluster water zone as allowed for deployment, which is capable of clustering any alternative so that the blue sections sensed as water covered zones and defined as allowed zone for vehicle deployment. The clustered map is presented by Fig.1.

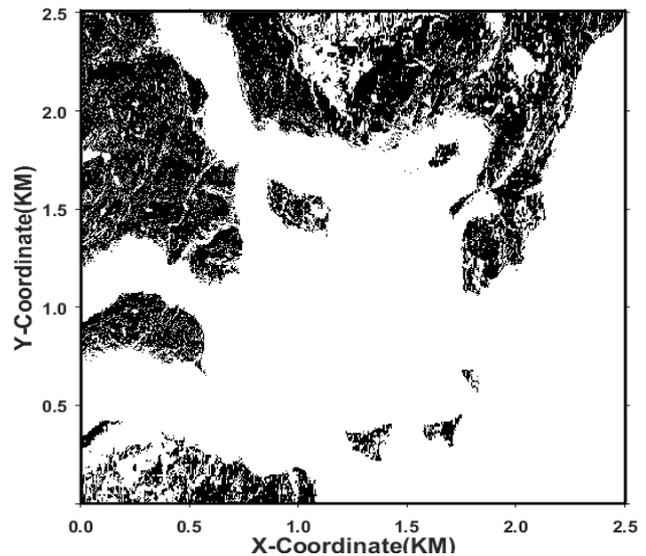

**Fig.1.** The clustered map of the for the path planner in which all area clustered to forbidden (black) and allowed (white) zones for vehicle's deployment.

In addition to offline map, in which the coast and rocks are considered forbidden edges, the uncertain static and moving obstacles ($\Theta_{(i)}$) are taken into account in which their radius, position and uncertainty modelled with normal distributions. The static uncertain obstacles has modelled with an uncertain radius varied iteratively in a specified bound with a Gaussian normal distribution of $\sim N(0,20)$ radiating oute from center of the object in a circular format. Another type is the uncertain moving obstacles that move to a random direction and are affected by current force, here the effect of current presented by uncertainty propagation proportional to current magnitude $U_R^C=|V_C|$. In both cases, obstacle's position and quantity initialized randomly with a Gaussian distributions in operating zone bundled to location of start and destination that vehicle is required to travel. An advanced path planning is able to generate energy efficient trajectories by using desirable current flow and avoiding undesirable current while avoid colliding obstacles. Desirable current can propel the vehicle on its trajectory and leads saving more energy. The state variables corresponding to NED and Body frame are shown in Eq.(1); then, the kinematics of the vehicle is described by a set of ordinary differential equations as shown in Eq.(2):

$$\eta : (X, Y, Z, \phi, \theta, \psi)$$
$$\upsilon : (u, v, w, p, q, r) \qquad (1)$$

$$\begin{bmatrix} \dot{X} \\ \dot{Y} \\ \dot{Z} \end{bmatrix} = [{}^n_b R] \begin{bmatrix} u \\ v \\ w \end{bmatrix}; \quad [{}^n_b R] = \begin{bmatrix} \cos\psi\cos\theta & -\sin\psi & \cos\psi\sin\theta \\ \sin\psi\cos\theta & \cos\psi & \sin\psi\sin\theta \\ -\sin\theta & 0 & \cos\theta \end{bmatrix}, \qquad (2)$$

where $X, Y,$ and $Z$, are the position of vehicle in North, East, Down (NED or $\{n\}$ frame), in turn; and $\phi, \theta, \psi$ are the Euler angles of roll, pitch, and yaw, respectively. The $u, v$ and $w$ are surge, sway and heave water referenced velocities $\upsilon$ of the body frame $\{b\}$, respectively. The $[{}^n_b R]$ is a rotation matrix that transforms the $\{b\}$ into the $\{n\}$ frame using Euler angles. It is assumed the vehicle moves with constant thrust power which means the vehicle moves with constant water-referenced velocity $\upsilon$. Applying the water current components of $V_C=(u_c, v_c)$, the surge $(u)$, sway$(v)$, and have$(w)$ components of the vehicle's velocity is calculated from the Eq.(3).

$$u = |\upsilon|\cos\theta\cos\psi + |V_C|\cos\theta_C\cos\psi_C$$
$$v = |\upsilon|\cos\theta\sin\psi + |V_C|\cos\theta_C\sin\psi_C \qquad (3)$$
$$w = |\upsilon|\sin\theta$$

here, $V_C$ is magnitude of current, $\psi_c$ and $\theta_c$ are directions of current in horizontal and vertical plans, respectively.

The potential trajectory $\wp_i$ in this research is generated based on B-Spline curves captured from a set of control points $\vartheta : \{\vartheta^1_{x,y,z}, \dots, \vartheta^i_{x,y,z}, \dots, \vartheta^n_{x,y,z}\}$, in which all $\vartheta$ should be located in respective search region constraint to predefined upper and lower bounds of $\vartheta \in [U^i_\vartheta, L^i_\vartheta]$ as presented by Eq.(4) and (5).

$$\wp^i_{x,y,z} = \sum_{i=1}^{|\wp|} \sqrt{(\vartheta^{i+1}_x - \vartheta^i_x)^2 + (\vartheta^{i+1}_y - \vartheta^i_y)^2 + (\vartheta^{i+1}_z - \vartheta^i_z)^2}$$
$$\wp(t) = [X(t), Y(t), Z(t), \psi(t), \theta(t), u(t), v(t), w(t)] \qquad (4)$$
$$\left. \begin{aligned} X(t) &= \sum_{i=1}^{n} \vartheta^i_x(t) \times B_{i,K}(t) \\ Y(t) &= \sum_{i=1}^{n} \vartheta^i_y(t) \times B_{i,K}(t) \\ Z(t) &= \sum_{i=1}^{n} \vartheta^i_z(t) \times B_{i,K}(t) \end{aligned} \right\}$$

$$\psi = \tan^{-1}\left(\frac{|\vartheta^{i+1}_y - \vartheta^i_y|}{|\vartheta^{i+1}_x - \vartheta^i_x|}\right)$$
$$\theta = \tan^{-1}\left(\frac{-|\vartheta^{i+1}_z - \vartheta^i_z|}{\sqrt{(\vartheta^{i+1}_y - \vartheta^i_y)^2 + (\vartheta^{i+1}_x - \vartheta^i_x)^2}}\right) \qquad (5)$$

where $B_{i,K}(t)$ is the curve's blending functions, $t$ is the time step, and $K$ is the order of the curve represents the smoothness of the curve, where bigger $K$ correspond to smoother curves. $X(t), Y(t), Z(t)$ display the vehicle's position at each time step along the generated path $\wp$ in the $\{n\}$ frame. Appropriate adjustment of the control points play a substantial role in optimality of the generated trajectory.

### 3 Structure of the DE-Based Time-Optimal Path Planning

The Differential Evolution (DE) algorithm (Price and Storn 1997) is a population-based algorithm that introduced as an improved version of the genetic algorithm and uses similar operators of selection, mutation, and crossover. The DE obtains better solutions and constructs faster optimization due to use of real coding of floating point numbers in presenting problem parameters. The algorithm has a simple structure and mostly relies on differential mutation operation and non-uniform crossover. The selection operator converges the solutions toward the desirable regions in the search space. The crossover operator shuffles the existing population members and searches for a better solution space. The offspring's are inherited in unequal proportions from the previous solution vectors. The process of the DE algorithm is given below:

### A. Initialization

In first step, an initial population $DE_{pop}$ of solution vectors $\chi_i$, ($i=1,...,nPop$) is randomly generated with uniform probability. For path planning purpose, any arbitrary path $\wp^i_{x,y,z}$ is assigned with solution vector $\chi_i^{x,y,z}$ where control points $\vartheta$ along the path $\wp^i_{x,y,z}$ correspond to elements of the solution vector. The solution space efficiently gets improved in each iteration $G$ applying evolution operators. A candidate solution vector is designated as

$$\begin{matrix} \forall i=1,...,nPop \\ \forall \chi_{i,G} \\ \forall G \in \{1,...,Iter_{max}\} \end{matrix} \Rightarrow \exists \begin{cases} \chi^x_{i,G} = \vartheta^i_x \\ \chi^y_{i,G} = \vartheta^i_y \\ \chi^z_{i,G} = \vartheta^i_z \end{cases} \tag{6}$$

where, $nPop$ is the number of individuals in DE population, $Iter_{max}$ is the maximum number of iterations.

### B. Mutation

The effectual modification of the mutation scheme is the main idea behind impressive performance of the DE algorithm, in which a weighted difference vector between two population members to a third one is added to mutation process that is called donor. Three different individuals of $\chi_{r1,G}$, $\chi_{r2,G}$, and $\chi_{r2,G}$ are selected randomly from the same generation $G$, which one of this triplet is randomly selected as the donor. So, the mutant solution vector is produced by

$$\begin{aligned} \dot{\chi}_{i,G} &= \chi_{r3,G} + F(\chi_{r1,G} - \chi_{r2,G}) \\ r1,r2,r3 &\in \{1,...,nPop\}, r1 \neq r2 \neq r2 \neq i, \quad F \in [0,1+] \end{aligned} \tag{7}$$

where $F$ is a scaling factor that controls the amplification of the difference vector ($\chi_{r1,G} - \chi_{r2,G}$). Giving higher value to $F$ promotes the exploration capability of the algorithm. The proper donor accelerates convergence rate that in this approach is determined randomly with uniform distribution as follows.

$$donor = \sum_{i=1}^{3} \left( \lambda_i \bigg/ \sum_{j=1}^{3} \lambda_j \right) \chi_{ri,G}, \tag{8}$$

where $\lambda_j \in [0,1]$ is a uniformly distributed value. This scheme provides a better distribution of the solution vectors. The mutant individual $\dot{\chi}_{i,G}$ and parent individual $\chi_{i,G}$ are then shifted to the crossover (*Recombination*) operation.

### C. Crossover

The parent vector to this operator is a mixture of individual $\chi_{i,G}$ from the initial population and the mutant individual $\dot{\chi}_{i,G}$. The produced offspring $\ddot{\chi}_{i,G}$ from the crossover is described by

$$\begin{aligned} \chi_{i,G} &= (x_{1,i,G},...,x_{n,i,G}) \\ \dot{\chi}_{i,G} &= (\dot{x}_{1,i,G},...,\dot{x}_{n,i,G}) \\ \ddot{\chi}_{i,G} &= (\ddot{x}_{1,i,G},...,\ddot{x}_{n,i,G}) \\ j &= 1,...,n; \quad n \in [1,nPop] \\ \ddot{x}_{j,i,G} &= \begin{cases} \dot{x}_{j,i,G} & rand_j \leq C_r \vee j = k \\ x_{j,i,G} & rand_j \leq C_r \wedge j \neq k \end{cases} \end{aligned} \tag{9}$$

where $k \in \{1,..., nPop\}$ is a random index chosen once for all population $nPop$. The second DE control parameter is the crossover factor $C_r \in [0,1]$ that is set by the user.

### D. Evaluation and Selection

The offspring produced by the crossover and mutation operations is evaluated. The best-fitted solutions produced by evolution operators are selected and transferred to the next generation ($G+1$).

$$\begin{aligned} \dot{\chi}_{i,G+1} &= \begin{cases} \dot{\chi}_{i,G} & f(\dot{\chi}_{i,G}) \leq f(\chi_{i,G}) \\ \chi_{i,G} & f(\dot{\chi}_{i,G}) > f(\chi_{i,G}) \end{cases} \\ \ddot{\chi}_{i,G+1} &= \begin{cases} \ddot{\chi}_{i,G} & f(\ddot{\chi}_{i,G}) \leq f(\chi_{i,G}) \\ \chi_{i,G} & f(\ddot{\chi}_{i,G}) > f(\chi_{i,G}) \end{cases} \end{aligned} \tag{10}$$

The performance of the offspring and parents are compared for each operator according to defined cost function $f$ and the worst individuals eliminated from the population.

## 3.1 Path Optimization Criterion

The path planning is an optimization problem in which the main goal is to minimize the travel time and distance while avoid colliding obstacles or coastal forbidden edges. Assuming the constant water referenced velocity $|v|$ for the vehicle, the travel time calculated by

$$\forall \wp_{x,y,z} \Rightarrow T_\wp = \sum_1^n t_i = \sum_1^n \frac{|\wp| \left| \vartheta_{i+1}^{t_{i+1}} - \vartheta_i^{t_i} \right|}{|v|} \tag{11}$$

where the $T_\wp$ is the path flight time. The resultant path should be safe and feasible to environmental constraints associated with the forbidden zones of map and avoidance of colliding obstacles, and coping with the current flow. Additional to environmental constraints, the vehicle kinematic restrictions on surge, sway yaw and pitch components and limitation on depth maneuver also should be taken into account. Hence, the path cost function is defined as a combination of the performance index, which is path time here, and corresponding penalty function as described in (12).

$$\nabla_{\Sigma M,\Theta} = \begin{cases} 1 & \wp_{x.y.z}(t) = Coast : Map(x,y) = 1 \\ 1 & \wp_{x.y.z}(t) \cap \bigcup_{N\Theta} \Theta \\ 0 & Otherwise \end{cases}$$

$$\nabla_\wp = \begin{cases} \varepsilon_{z\min} \times \min(0; Z(t) - Z_{\min}) \\ \varepsilon_{z\max} \times \max(0; Z(t) - Z_{\max}) \\ \varepsilon_u \times \max(0; u(t) - u_{\max}) \\ \varepsilon_v \times \max(0; |v(t)| - v_{\max}) \\ \varepsilon_\theta \times \max(0; \theta(t) - \theta_{\max}) \\ \varepsilon_\psi \times \max(0; |\dot{\psi}(t)| - \psi_{\max}) \\ \varepsilon_{\Sigma M,\Theta} \times \nabla_{\Sigma M,\Theta} \end{cases} \tag{12}$$

$$C_\wp = T_\wp + \sum_{i=1}^n Q_i f(\nabla_\wp)$$

where, $\Theta$ represents any arbitrary obstacle, $Q_i f(\nabla_\wp)$ is a weighted violation function that respects the AUV kinodynamic and collision constraints including depth violation ($Z$), to prevent the path from straying outside the vertical operating borders, surge ($u$), sway ($v$), yaw ($\psi$), pitch($\theta$) violations, and the collision violation ($\nabla_{\Sigma M,\Theta}$) specified to prevent the path from collision danger. The $\varepsilon_{zmin}$, $\varepsilon_{zmax}$, $\varepsilon_u$, $\varepsilon_v$, $\varepsilon_\theta$, $\varepsilon_\psi$, and $\varepsilon_{\Sigma M,\Theta}$ respectively denote the impact of each constraint violation in calculation of total path violation, and $C_\wp$ is the path cost.

## 4 Discussion on Simulation Results

Current research investigates the performance of DE algorithm on AUV path planning problem in uncertain and partially known underwater environment in different situations and aims to satisfy performance aspects of "total path length/time", "vehicle kinematic constraints" and "collision avoidance" for evaluation of an optimal path. The vehicles water-referenced velocity has been set on 3 m/s. Population size is set in 100, the maximum number of iterations is set on 200, lower and upper bound of scaling factor is set on 0.2 and 0.8, respectively. The crossover probability fixed on 20 percent. Number of control points for each B-spline path set on 5.

Operating scale is assumed to be three dimensional terrain $\Gamma_{3D}$ of 2.5 $km^2$ and depth of 1000 $m$. Environmental risks and factors such as static ocean current map and uncertain static/moving obstacles are considered to promote models realistic features and motion performance. The efficient path planner is able to use desirable current flow to save the energy and must be capable of adapting undesirable current force. Fig.2 represents the performance of the introduced DE-based path planner in adapting ocean current. Referring to Fig.2, it is apparent that the produced path adapts the current behaviour as the path curve is deformed according to current arrows. The colormap represents the current magnitude that is calculated according to correlations of current vectors, where the red sections correspond to higher intensity of current flow and as the colour gets lighter toward blue, the intensity of the current magnitude gets reduced.

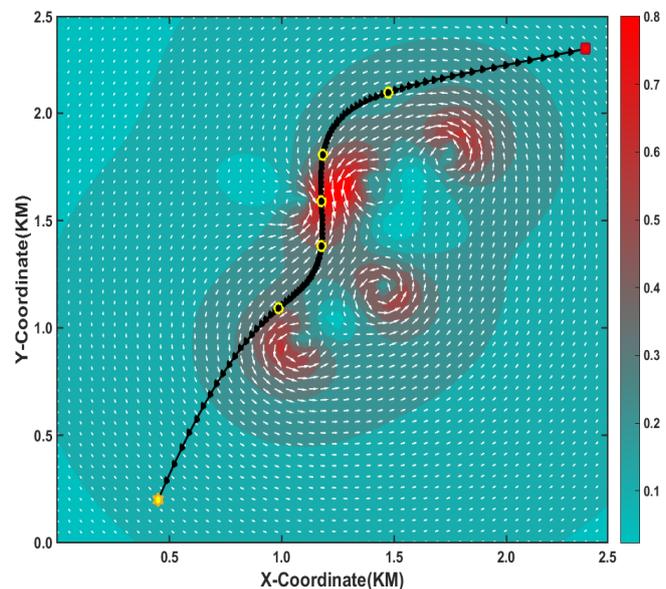

**Fig.2.** The path deformation in adapting current flow

The efficiency of the produced trajectory is examined encountering map data, current data, and static and moving obstacles generated randomly and configured individually with random position by the mean variance of their uncertainty distribution and velocity proportional to current velocity presented by Fig.3. The obstacles collision boundary grows gradually with time at rates randomly selected based on type of the obstacle with or without encountering current effect (velocity). The goal is generating a shortest collision-free trajectory from the starting point to the destination position. A three dimensional plot is used to present the path performance more clear given by Fig.4. If vehicles trajectory does not cross inside the corresponding collision boundary, no collision will occur. The coastal areas in Fig.3 is also forbidden for vehicle's deployment. In simulation result given by Fig.3 and Fig.4, the obstacles dimension and direction changes by time in vehicles operation window.

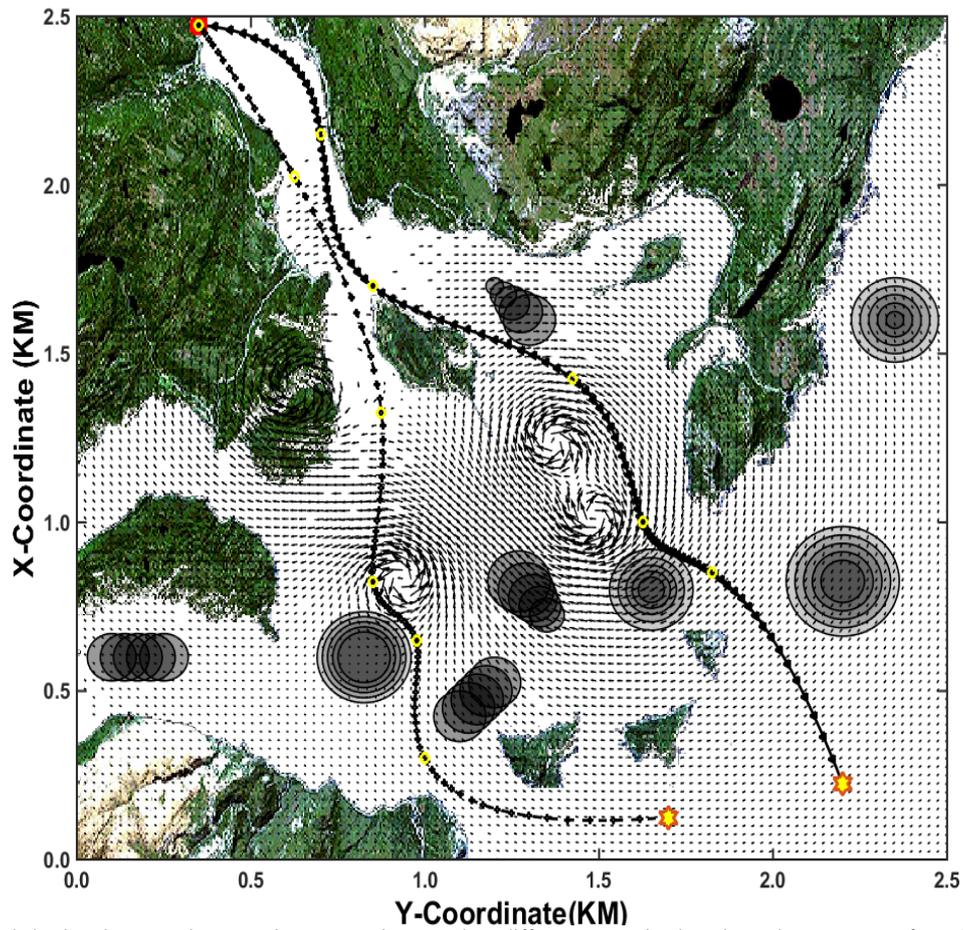

**Fig.3.** The path behaviour in a complex scenario encountering map data, different uncertain obstacles and water current from the same start point two different destinations

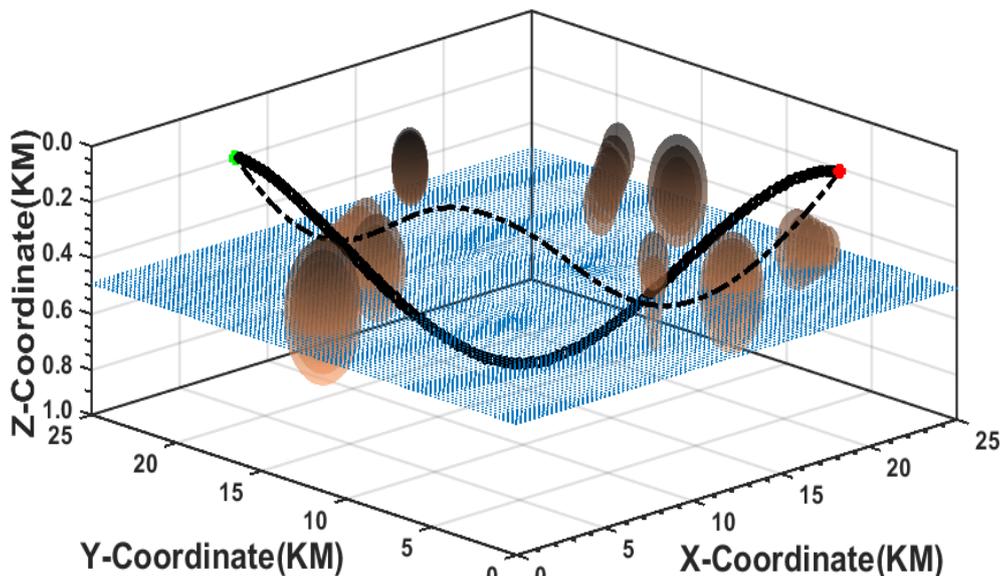

**Fig.4.** Deformation the optimal path and its alternative in three-dimension space

Increasing the complexity of the obstacles, increases the problems complexity, however, it is derived from simulation results in Fig.3 and Fig.4 that the proposed path planner is able to accurately handle the collision avoidance operation regardless of type or number of appeared obstacles, while taking the efficient path in accordance with current vectors. It is noted from Fig.5 that the violation value gradually diminishes as the algorithm iterates, also considering the cost variations it is noted the algorithm experiences a moderate convergence by passing iterations as the cost decreases iteratively, which declares that algorithm enforces the solutions to approach the optimum answer (path) with minimum cost and efficiently manages the path to eliminate the collision penalty within 100 iterations.

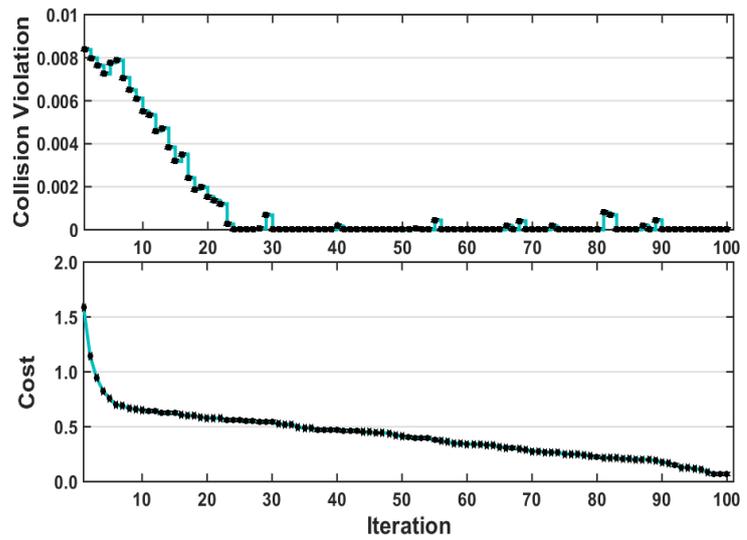

**Fig.5.** Variations of cost and collision violation for DE-based path planning over 100 iterations

In order to compute the B-spline curve, the angular velocity is required at all times. The vehicle kinematic constraints are calculated based on the paths' curvature performed by the spline that corresponds to vehicles position along the path.

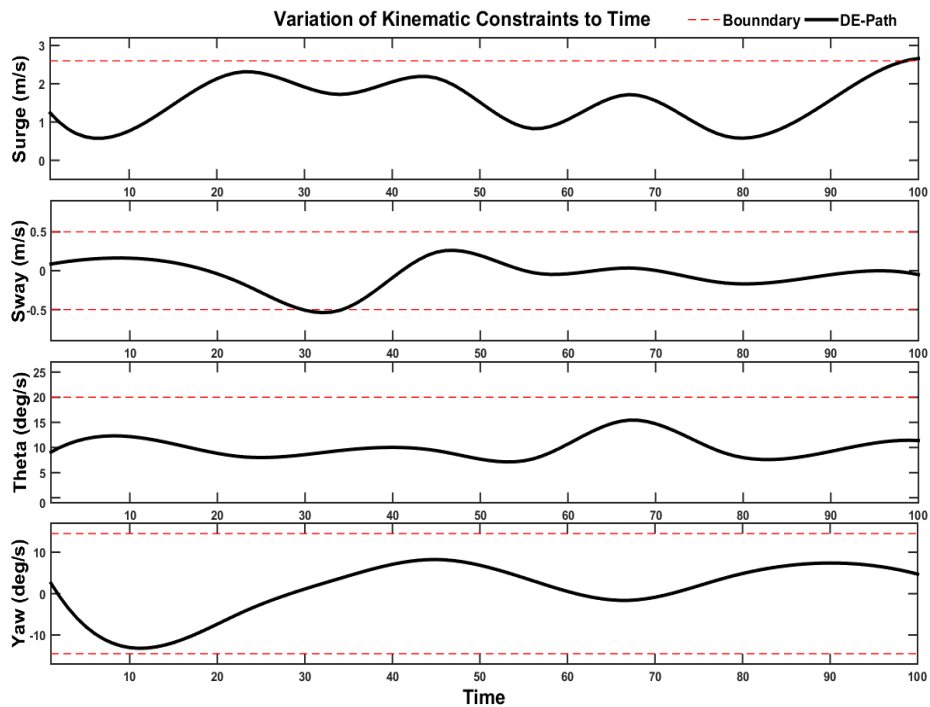

**Fig.6.** Variations of vehicle's kinematic constraint for B-spline curves generated by DE based path planner over the time

As presented in Fig.6 almost all constraint are satisfied and the yaw, surge, sway, and theta vary in the specified range (presented by red dashed line) without any violation. Altogether, it is derived from simulation results that the algorithm is able to accurately handle the collision avoidance, satisfy problem constraints, and tend to find optimum path regardless of complexity of the terrain.

## 5    Conclusion

Authentic and efficient path planning is a critical primary requirements to mission success and safe operation of an AUV in uncertain unknown ocean environments. Many research carried out on autonomous vehicles path/trajectory planning over the past decades suggesting variety of evolutionary or swarm based optimization algorithms. Undoubtedly there is always a significant requirement for more efficient techniques in autonomous vehicle's path planning problem and promoting modelling of the environment to be more analogous to real world ocean terrain.

The primary step toward increasing endurance and range of vehicle's operation is promoting vehicles capability in saving more energy. Underwater environment includes turbulent water currents and various dynamic/static obstacles, which usually may not be fully known at the beginning of the mission. Obstacles may be appeared or change behavior as the vehicle moves through the environment. The vehicle should be autonomous enough to cope dynamic changes over the time while it is resistance against various disturbances. Advanced path planning is able to generate energy efficient trajectories by using desirable current flow and avoiding undesirable current. Desirable current can propel the vehicle on its trajectory and leads saving more energy. This issue considerably diminish the total AUV mission costs. To fulfil the objectives of this research toward solving the stated problems associated with AUV path planning approaches in such a severe unknown environment, this research applied Differential Evolution algorithm and the terrain is modelled in a realistic feature to cover possibilities of the real world terrain. In this respect, efficiency of the proposed DE-based path planner is investigated encountering real map data, uncertain static/moving obstacles and static water current map captured from superposition of multiple viscous Lamb vortices and 2-D Navier-Stokes equation (Garau et al. 2006). It is noteworthy to mention from analyze of the simulation results that the proposed approach is able to generate trajectory resistant to terrain deformation and respects environmental and kinematic constraints. The proposed solutions accurately cope with adverse currents as well as exploits desirable currents to enhance the operation speed that leads significant saving of energy and furnishes the expectation of the vehicle at higher level of the autonomy for real-time implementation.

For future researches, the modelling of the environment will be promoted to be more realistic. Also internal situations and vehicles fault tolerance also will be encountered to increase vehicle total situational awareness and self-awareness toward facilitating the vehicle in robust decision making accordingly.